\documentclass{article}

    \PassOptionsToPackage{numbers, compress}{natbib}

\usepackage[final]{neurips_2023}

\usepackage[utf8]{inputenc} 
\usepackage[T1]{fontenc}    
\usepackage{hyperref}       
\usepackage{url}            
\usepackage{booktabs}       
\usepackage{wrapfig}
\usepackage{amsfonts}       
\usepackage{nicefrac}       
\usepackage{microtype}      
\usepackage{xcolor}         
\usepackage{multirow}
\usepackage{graphicx}
\usepackage{mathtools}
\usepackage{amsmath}
\usepackage{amsthm}
\usepackage{adjustbox}
\usepackage{algorithm}
\usepackage{algpseudocode}
\usepackage[noend]{algcompatible}
\usepackage{algorithmicx}

\usepackage{paralist}
\let\itemize\compactitem
\let\enditemize\endcompactitem
\let\enumerate\compactenum
\let\endenumerate\endcompactenum

\theoremstyle{plain}
\newtheorem{theorem}{Theorem}[section]

\theoremstyle{definition}

\newtheorem{remark}[theorem]{Remark}

\DeclareMathOperator*{\minimize}{minimize \ }

\algnewcommand\algorithmicreturn{\textbf{return}}
\algnewcommand\RETURN{\State \algorithmicreturn}%

\algrenewcommand\algorithmicrequire{\textbf{function}} 

\algnewcommand\algorithmicreq{\textbf{Require:} }
\algnewcommand\REQ{\STATEx \algorithmicreq{}}%

\usepackage[capitalize]{cleveref}
\crefname{section}{Sec.}{Secs.}
\Crefname{section}{Section}{Sections}
\Crefname{table}{Table}{Tables}
\crefname{table}{Tab.}{Tabs.}

\title{Spectral Co-Distillation for Personalized Federated Learning}

\author{
   Zihan Chen$^1$, Howard H. Yang$^2$, Tony Q.S. Quek$^1$, and Kai Fong Ernest Chong$^1$\thanks{Corresponding author} \\
   $^1$Singapore University of Technology and Design (SUTD) \\
   $^2$Zhejiang University/University of Illinois Urbana-Champaign Institute, Zhejiang University \\
   \texttt{zihan\_chen@sutd.edu.sg}
}

\begin{document}

\maketitle

\begin{abstract}
Personalized federated learning (PFL) has been widely investigated to address the challenge of data heterogeneity, especially when a single generic model is inadequate 
in satisfying the diverse performance requirements of local clients simultaneously. Existing PFL methods are inherently based on the idea that the relations between the generic global and personalized local models are captured by the similarity of model weights. Such a similarity is primarily based on either partitioning the model architecture into generic versus personalized components, or modeling client relationships via model weights. To better capture similar (yet distinct) generic versus personalized model representations, we propose \textit{spectral distillation}, a novel distillation method based on model spectrum information. Building upon spectral distillation, we also introduce a co-distillation framework that establishes a two-way bridge between generic and personalized model training. Moreover, to utilize the local idle time in conventional PFL, we propose a wait-free local training protocol. Through extensive experiments on multiple datasets over diverse heterogeneous data settings, we demonstrate the outperformance and efficacy of our proposed spectral co-distillation method, as well as our wait-free training protocol.
\end{abstract}

\section{Introduction}\label{sec:intro}
With the rapid rise in mainstream popularity of artificial intelligence (AI) models such as ChatGPT~\cite{van2023chatgpt} and LoRA~\cite{hu2021lora}, there has been an increasing shift towards the development of personalized AI assistants~\cite{tan2022towards}. 
Hence, in a future where personalized AI services become mainstream, training AI models on personal data while preserving data privacy would become increasingly important~\cite{albrecht2016gdpr}, and maintaining the quality for such models would require collaborative training across multiple models. 
Personalized federated learning (PFL) emerges as a promising privacy-preserving distributed learning paradigm that is well-equipped to meet such requirements~\cite{tan2022towards}. 
As an extension of federated learning (FL), PFL aims to train a customized machine learning model for each client or each group of clients with similar preferences~\cite{deng2020adaptive}. 
When faced with inconsistencies in the objective functions of different clients, conventional FL fails to generalize well with just a single model, while in contrast PFL promises to generalize well across all clients, even in the presence of data heterogeneity (e.g., label distribution skew and label quantity skew)~\cite{wang2021field, xu2022fedcorr, yang2019survey,hsu2019measuring, li2020prox, acar2020federated,chen2021dynamic, TangYLT22, chen2022towards}.

To tackle the challenges of personalization, numerous works have focused on designing new PFL systems, or enhancing the performance of personalized models from different aspects, such robustness, fairness, and model convergence\cite{li2021ditto,li2021fedmask,t2020moreau,chen2022pflbench}.
Under federated settings, personalization is achieved through capturing the (dis-)similarity of the local versus globally shared model representations. 
In practical FL/PFL applications of collaboratively training deep neural networks (DNNs), only the model parameters (e.g., model weights or gradients) are exchanged between the clients and the server~\cite{yang2019survey,mcmahan2017communication}.
Existing DNN-based PFL methods capture this (dis-)similarity either by decoupling the model architecture into groups of layers/channels\cite{liang2020think,shen2022cd2,arivazhagan2019federated,chen2022on,collins2021exploiting,pillutla2022partial}, or by designing local optimization methods with regularization based directly on model weights~\cite{t2020moreau,li2021ditto}. 
Unfortunately, the motivations for such approaches are based on empirical observations, without an overarching theory to explain model (dis-)similarity in relation to training dynamics.
 
In deep learning theory, the training dynamics of DNNs have been studied from the lens of Fourier analysis~\cite{xu2022overview}.
A crucial insight from this analysis is that there is an implicit self-regularization effect arising from the training process itself. 
Given a target function $f$ to learn, the model tends to learn the lower frequencies of the Fourier spectrum of $f$ first before learning the respective higher frequencies.  
Such a bias in this training process is called \textit{spectral bias}~\cite{rahaman2019onspectral,fridovich2022spectralpractice}.
Informally, spectral bias describes the commonly encountered phenomenon that DNNs first learn low-level features before learning high-level features. 

Motivated by this insight, we can distinguish different levels of features in a model representation by looking at its Fourier spectrum.  
Intuitively, diverse personalized models would still share the same low-level features, and a global generic model would contain the same low-level features. 
Hence, despite any inconsistencies in the objective functions of different clients, there would be no conflict in learning low-level features for both the generic and personalized models.
Consequently, with the expected similarity in the lower frequency components of the Fourier spectra of both the generic and personalized models, we can distill the knowledge of the lower Fourier coefficients to boost the performance of the generic model. Dually, the entire Fourier spectrum of the generic model, which includes the ``averaged'' high-level features across all clients, would benefit the training of the personalized models.
By combining both perspectives, \textit{we shall propose a co-distillation framework for PFL that captures (dis-)similarity in models via spectral information.}

Typically, when designing PFL systems, a compute-and-wait protocol is implicitly assumed for local training~\cite{li2021ditto,chen2022on}. %
This means that the locally updated generic models would be sent by the clients to the server after all local computation tasks have been completed.
Such a protocol would yield a period of idle waiting where clients have to wait for the next aggregated model to be broadcasted. 
\textit{By circumventing this compute-and-wait protocol, we shall utilize the local idle time for training to reduce the total PFL runtime.}
 
Overall, our contributions can be summarized as follows:
\begin{itemize}
    \item We propose a spectral co-distillation framework for PFL. In particular, this is the first ever use of spectral distillation in PFL to capture the (dis-)similarity of the generic and personalized models. Also, this is the first ever bi-directional knowledge distillation directly between the generic and personalized models.
    \item 
    We propose a wait-free local training protocol for our spectral co-distillation framework, where we utilize the idle time during global communication so as to reduce the total PFL runtime.
    \item Through extensive experiments on multiple datasets with heterogeneous data settings, we demonstrate the outperformance and efficacy of our proposed spectral co-distillation framework with the wait-free communication protocol design for PFL, with respect to model generalizability and the total PFL runtime.
\end{itemize}

\section{Related work}\label{sec:related}
\textbf{PFL.} In PFL, prior efforts have focused on training multiple personalized models via leveraging the similarity and relationships between the global generic model and the local personalized models, such as via model interpolation/mixture~\cite{hanzely2020federated}, model decoupling~\cite{arivazhagan2019federated}, and personalized optimization with customized regularizers~\cite{li2021ditto}. 
In DNN-based FL applications, decoupling-based approaches divide the model into a private part (kept at the local side) and a shared part (exchanged between the server and clients)~\cite{tan2022towards,pillutla2022partial,chen2022on}. In particular, FedPer~\cite{arivazhagan2019federated} and FedRep~\cite{collins2021exploiting} share the shallow layers and train personalized deep layers, while in contrast, LG-Fed~\cite{liang2020think} and CD$^2$-pFed maintain personalized shallow layers and channels~\cite{shen2022cd2}, respectively. 
Moreover, Fed-RoD proposes a framework to achieve state-of-the-art (SOTA) performance for generic and personalized models simultaneously, based on the ``two-loss, two-predictor'' design\cite{chen2022on}.
APFL~\cite{deng2020adaptive} and L2GD~\cite{hanzely2020federated} consider using a mixture of local and global models to achieve personalization, in which the mixture weight controls the personalization level. 
Personalized local training methods have been recently explored, which include local fine-tuning in FedBABU~\cite{oh2022fedbabu}, bi-level optimization in Ditto~\cite{li2021ditto}, feature alignment in FedPAC~\cite{xu2023fedpac}, and personalized model sparsification in FedMask~\cite{li2021fedmask,li2021lotteryfl} and PerFedMask~\cite{setayesh2023perfedmask}.
More broadly, meta-learning~\cite{jiang2019improving,fallah2020pfl_maml}, gaussian processes~\cite{achituve2021gpkernel}, and hyper-network-based approaches~\cite{shamsian2021hypernet} have been investigated in PFL.
Specifically, there is another type of PFL that aims to train personalized models at the level of clusters of clients with similar preferences~\cite{sattler2020clustered,duan2021fedgroup,ghosh2020efficient}.

\textbf{Knowledge Distillation (KD) in FL.} KD has been widely explored in knowledge transfer scenarios, which usually is used to transfer knowledge from the pre-trained teacher model to the student model via minimizing the distance from the latent or logit outputs of the two models~\cite{chen2019knowledge,qian2022switchable}. KD-based FL frameworks have been developed with diverse setups, such as FedMD~\cite{li2019fedmd} and FedDF~\cite{lin2020ensemble}.
On the other hand, knowledge-transfer-based PFL frameworks are investigated in \cite{cho2023communication,he2020group} with different model structures at the local clients, which could address the system heterogeneity and improve communication efficiency. However, such methods rely on the assumption of having access to a public labeled/unlabelled dataset, which may not be a realistic assumption in FL applications~\cite{tan2022towards}.
Moreover, co-distillation methods have been investigated in communication-efficient decentralized scenarios to improve generalizability~\cite{cho2023communication}.

\begin{figure*}[t!]
\centering\includegraphics[width=1.0\textwidth]{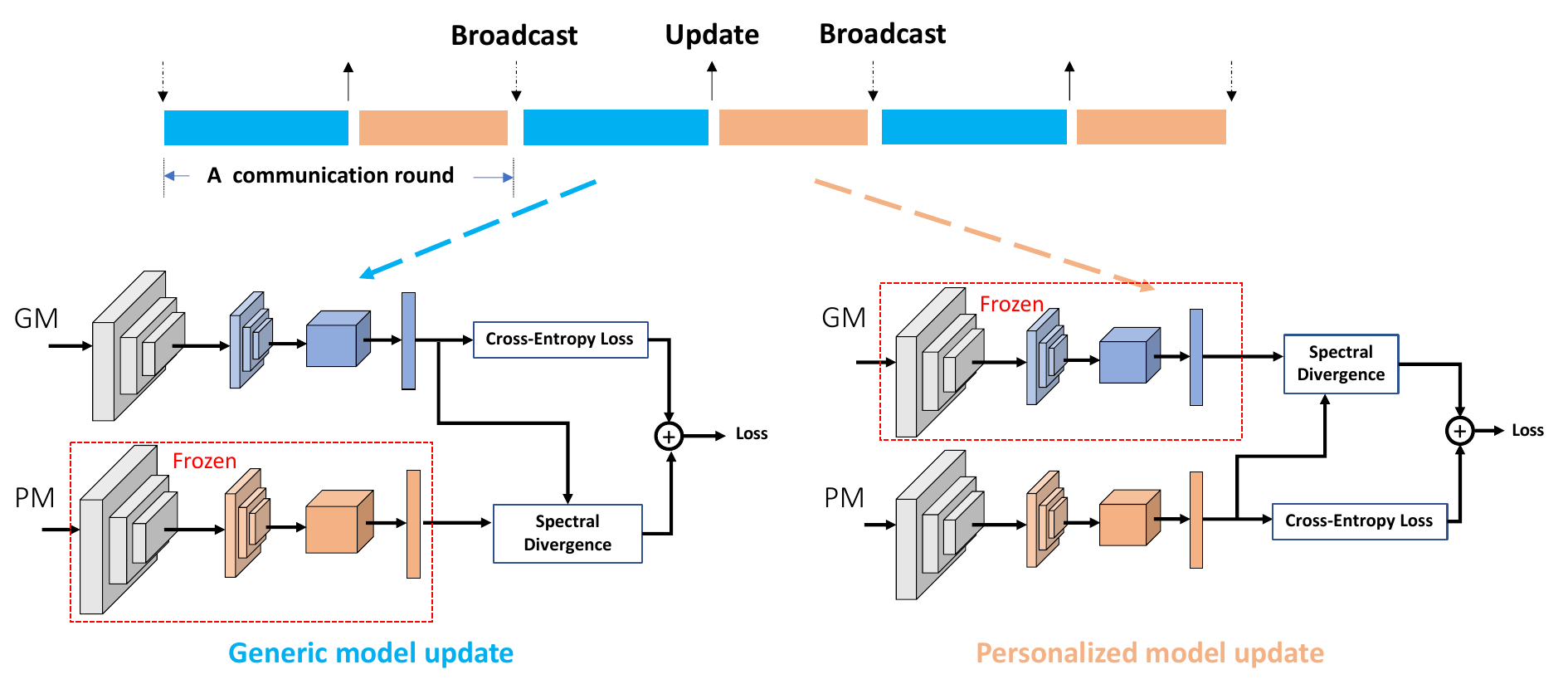} 
\caption{Spectral co-distillation framework with wait-free local training for PFL, in which the generic model (GM) training and the personalized model (PM) training are carried out via spectral distillation in two different stages.}
\label{fig:overview}
\end{figure*}

\section{Proposed framework}
\label{sec:propose_fw}
The main goal of this work is to train a generic global model and multiple personalized models simultaneously.
As summarized in Sec.~\ref{sec:intro}, our proposed framework consists of three major components: spectral distillation-based personalized model training, spectral co-distillation-based generic model training, and the wait-free sequential computation-communication protocol.
In this section, we first provide the preliminary and problem formulation for PFL and model spectrum in Sec.~\ref{subsec:pre}. Next, we present our proposed spectral distillation approach for PFL in Sec.~\ref{subsec:pm_train}, co-distillation-based generic model training in Sec.~\ref{subsec:gm_train}, and the wait-free local training protocol in Sec.~\ref{subsec:com_protocol}, accordingly. Moreover, the summarized algorithm is given in Sec.~\ref{subsec:algo}.
 
\subsection{Preliminaries}\label{subsec:pre}
\textbf{Problem formulation for FL and PFL.}
Consider an FL system consisting of a server and $N$ clients, in which client $i$ has a loss function $f_i:\mathbb{R}^d\rightarrow\mathbb{R}$ used for training on its local private dataset $\mathcal{D}_i=\{(x_i^j,y_i^j)\}_{j=1}^{n_i}$, where $n_i=|\mathcal{D}_i|$ denotes the size of the local dataset of client $i$. 
In conventional FL, the objective of all the participating clients in this system is to find a global model $w\in \mathbb{R}^d$ that solves the following minimization problem~\cite{mcmahan2017communication}: 
\begin{equation}\label{g1}
\minimize_{w \in \mathbb{R}^d} \left\{F(w):=\sum_{i=1}^N \frac{n_i}{n} f_i(w) 
 \right\},
\end{equation} 
where $n=\sum_{i=1}^Nn_i$ is the total number of training samples across the $N$ clients. 
In a typical communication round $t$, a subset $\mathcal{S}_t$ of clients is selected to conduct local training, starting from the latest global model weights $w_\text{G}^{t}$. 
Let $w_i^t$ denote the weights of client $i$'s model after local training. 
At the end of communication round $t$, the server would collect local models from the selected clients to update the global model via Federated Averaging (FedAvg), i.e. $w_\text{G}^{t+1} \gets \sum_{i \in \mathcal{S}_t} p_i^t w_i^t$, in which $p_i^t= n_i/\sum_{k \in \mathcal{S}_t}n_k$ represents the ratio of the local data samples in client $i$ over the total number of data samples in the selected subset $\mathcal{S}_t$ of clients for communication round $t$.

There are two general types of PFL: a) training $N$ personalized models for all $N$ clients; and b) training $1$ generic model and $N$ personalized models simultaneously.
In this work, we investigate the latter one, which we term as ``PFL+''. %
This means each client $i$ has a local personalized model $w_{\text{p},i}$ for its private dataset $\mathcal{D}_i$, and all clients jointly participate in the training of the generic model $w_\text{G}$.
After local training at client $i$, the updated generic model is denoted by $w_{\text{G},i}$.
Thus, PFL can be formulated using a regularized loss function with regularization term $R_\text{p}(w_{\text{p},i}, w_{\text{G},i})$.  
For example, $R_\text{p}(w_{\text{p},i}, w_{\text{G},i})$ could represent the similarity/divergence between the global and local models' features, such as model weights, feature centroids, and prototypes. 
In our method, $R_\text{p}(w_{\text{p},i}, w_{\text{G},i})$ represents cross-model distillation during the training of client $i$'s personalized model. 
Therefore, the objective of personalized model training in PFL+ can be formally formulated as a bi-level optimization problem~\cite{t2020moreau}:

\begin{align}\label{eq:pm_obj}
    \text{(\textbf{P1}):} \quad \minimize_{w_{\text{p},i}\in \mathbb{R}^d}&\quad \Big\{f_{\text{p},i}(w_{\text{p},i}):=  f_i(w_{\text{p},i}) + \lambda_\text{p} R_\text{p}(w_{\text{p},i}, w_{\text{G},i})\Big\} \text{ \quad for each client $i$}\\
    \text{subject to} & \quad w_{\text{G},i} \gets \text{ updated generic model from $w_\text{G}$,} 
\end{align} 

where the  regularization coefficient $\lambda_\text{p}$ is used to control the level of personalization.
For client $i$, when referring to a specific communication round $t$, we shall denote the untrained personalized model and updated generic model by $w_{\text{p},i}^{t-1}$ and $w_{\text{G},i}^{t}$, respectively.

\subsection{Personalized local model training}\label{subsec:pm_train} 
Motivated by both theoretical and empirical insights of the spectral bias inherent in the training dynamics of DNNs, we explore the use of the Fourier spectrum of the generic model for knowledge distillation to enhance the training of personalized local models.
In particular, we propose a distillation regularization term representing the divergence between the \textit{full} model spectra of the generic and personalized models. 

First, we introduce some notation.
Given vectors $p = (p_1, \dots, p_d)$, $q = (q_1, \dots, q_d)$ in $\mathbb{R}^d$, define the divergence function $\mathfrak{D}(p\|q) :=\sum_{i=1}^d p_i \log p_i - p_i \log q_i$. (By convention, $0\log 0 := 0$.) Note that when $p$ and $q$ are stochastic vectors representing parameter vectors of multinomial distributions $P$ and $Q$, then $\mathfrak{D}(p\|q)$ is identically the Kullback--Leibler (KL) divergence from $P$ to $Q$.
Next, let $\texttt{DFT}: \mathbb{C}^d \to \mathbb{C}^d$ denote discrete Fourier transform, let $\varrho: \mathbb{C}^d \to \mathbb{R}^d$ be the map given by $(z_1, \dots, z_d) \mapsto (\|z_1\|, \dots, \|z_d\|)$, and
define the function $s:\mathbb{R}^d \rightarrow \mathbb{R}^d$ by $s := \varrho \circ \texttt{DFT}$.
For an input vector of the weights of a DNN model, the output vector after applying $s$ shall be called the \textit{spectrum vector} of that model~\cite{fridovich2022spectralpractice}. 
Thus, in communication round $t$, the spectrum vectors of the personalized model $w_{\text{p},i}^{t-1}$ of client $i$ and updated generic model $w_{\text{G},i}^t$ are written as $s(w_\text{G}^t)$ and $s(w_{\text{p},i}^{t-1})$, respectively. 
We shall represent the divergence of the personalized and generic models by $ \mathfrak{D}(s(w_{\text{p},i}^{t-1}) \|s(w_{\text{G},i}^t))$, the divergence of their spectrum vectors.

Concretely, we define $R_\text{p}(w_{\text{p},i}, w_{\text{G},i}) :=  \mathfrak{D}(s(w_{\text{p},i}^{t-1}) \|s(w_{\text{G},i}^t))$, and let $f_i$ be the cross-entropy loss $\mathcal{L}_\text{CE}$ for all $i$.
Then the personalized objective function $f_{\text{p},i}$
of client $i$ in communication round $t$ (cf.~\eqref{eq:pm_obj}) is given by:
\begin{equation}\label{eq:4}
    \mathcal{L}^{\text{p}} (w_{\text{p},i}^{t-1}|w_{\text{G},i}^t) := \mathcal{L}_\text{CE}(w_{\text{p},i}^{t-1}| \mathcal{D}_i)+\lambda_\text{p}  \mathfrak{D}(s(w_{\text{p},i}^{t-1}) \|s(w_{\text{G},i}^t)).
\end{equation}
For simplicity, we use a common time-invariant $\lambda_\text{p}$ for all clients throughout training.
Since we are distilling the knowledge of the spectrum vector $s(w_{\text{G},i}^t)$ in \eqref{eq:4},
we term our approach as \textit{spectral distillation}.

\subsection{Generic model training}\label{subsec:gm_train}
Given a PFL+ training framework, it is natural to connect the \textit{roles of generic and personalized models} to the \textit{roles of the teacher and student models in distillation}, where the training of one model is guided by the knowledge distilled by the other. 
Co-distillation extends this idea. 
Intuitively, the role of each model alternates between teacher and student for knowledge distillation during training. 
In PFL+, since we are concurrently training both the generic and personalized models, either of them could be used for knowledge distillation.
The key challenge for applying co-distillation to PFL+ is that it is not obvious what knowledge should be distilled from the personalized models to enhance the training performance of the generic model.

In the theory of deep learning, it is well-known that when training a DNN, there is a learning bias towards the lower frequencies of its Fourier spectrum \cite{rahaman2019onspectral,fridovich2022spectralpractice}.
In fact, the lower-frequency components of this spectrum are robust to random weight perturbations.
Hence, with diverse personalized models, we would still expect the lower-frequency components of the spectra of all models (both generic and personalized) to be similar.
Consequently, we could use such lower-frequency components for knowledge distillation to enhance generic model training.

Motivated by this, we propose a truncated spectrum-based distillation loss as the regularizer for generic model training. 
Given $0<\tau\leq1$, let $\imath_{\tau}: \mathbb{R}^d \to \mathbb{R}^{\lceil \tau d\rceil}$ be the projection map onto the first $\lceil\tau d\rceil$ entries,
and define $\widehat{s} := \imath_{\tau}\circ s$.
Then the loss function for generic model training, which depends on the truncated spectrum vectors $\widehat{s}(w_{\text{G},i}^t)$ and $\widehat{s}(w_{\text{p},i}^{t-1})$, is given by:
\begin{equation}
    \mathcal{L}^{\text{G}} (w_{\text{G},i}^t|w_{\text{p},i}^{t-1}) := \mathcal{L}_\text{CE}(w_{\text{G},i}^t | \mathcal{D}_i)+\lambda_\text{g}  \mathfrak{D}(\widehat{s}(w_{\text{G},i}^t)\|\widehat{s}(w_{\text{p},i}^{t-1})),
\end{equation}
where the regularization term $R_\text{G}(w_{\text{G},i}^t, w_{\text{p},i}^{t-1}) := \mathfrak{D}(\widehat{s}(w_{\text{G},i}^t)\|\widehat{s}(w_{\text{p},i}^{t-1}))$ depends on the hyperparameter $\tau$, and $\lambda_\text{g}$ is the coefficient of this regularization term.
Analogous to \textbf{(P1)}, the objective of generic model training in PFL+ could be formulated as the following bi-level optimization problem:
\begin{align}\label{eq:gm_obj}
    \text{\textbf{(P2)}:} \quad \minimize_{w_\text{G}\in \mathbb{R}^d}&\quad \left\{f(w_\text{G}):=\sum_{i=1}^N \frac{n_i}{n} \left( f_i(w_\text{G}) + \lambda_\text{G} R_\text{G}(w_\text{G},w_{\text{p},i}) \right) \right\}\\
    \text{subject to} & \quad w_{\text{p},i} \gets \text{output of \textbf{(P1)} for client }i, \text{for } i=1,\dots, N.
\end{align}

Overall, by combining the two spectral distillation approaches together, we get a training framework for PFL+, which we shall call \textit{spectral co-distillation}.

\begin{figure*}[t!]
\centering\includegraphics[width=0.93\textwidth]{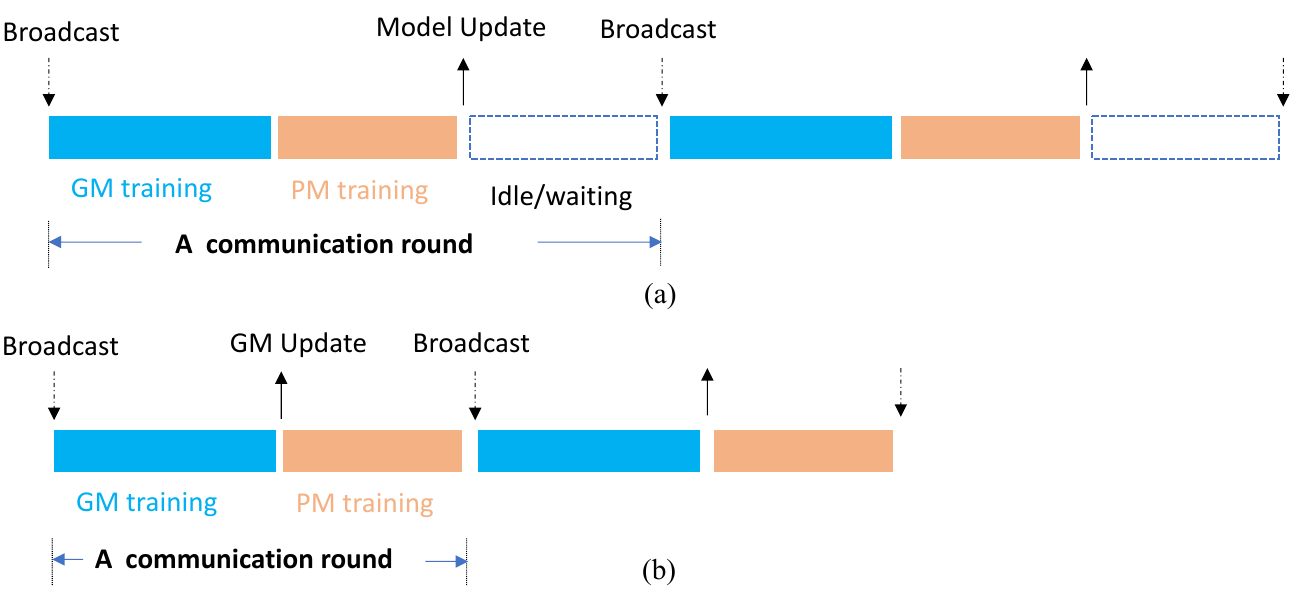} 
\caption{A comparison of the (a) conventional compute-and-wait protocol with the (b) proposed wait-free training protocol.}
\label{fig:protocol}
\end{figure*}

\subsection{Wait-free Local Training Protocol} \label{subsec:com_protocol}
In the context of federated computing, the total runtime, which includes both local computation and communication time throughout the entire training process, is a direct indicator of communication efficiency. However, current PFL frameworks adopt a compute-and-wait protocol for local training. This means that in each round, the client performs both generic and personalized model updates only after all local computation tasks have been completed, and resumes local training upon receiving the latest global model broadcasted from the server.
In consequence, there is idle waiting time between model update and model broadcast; see \cref{fig:protocol}(a).

To improve the communication efficiency of PFL training with respect to the total runtime, we propose a \textit{wait-free local training protocol}, as depicted in \cref{fig:protocol}(b). In our protocol, the client updates the generic model according to the conventional generic FL training and trains the personalized model during the global communication time period. 
Unlike existing PFL frameworks, local clients would send the updated generic model to the server before the start of the personalized model training.
Thus, our protocol eliminates idle waiting time, thereby dramatically reducing total runtime.
Furtherfore, it could be easily incorporated into existing PFL frameworks, such as Ditto~\cite{li2021ditto}, to further improve the efficiency; see \cref{tab:comm_cost}. 

\textbf{Discussion on the proposed protocol and related work.} Our proposed wait-free local training protocol is specially designed for the PFL+ scenario, where each client trains two models locally. 
For simplicity, we use this protocol in our experiments, under the assumption of synchronized PFL+. For comparison in the asynchronized PFL+ setting~\cite{nguyen2022buffered}, see Appendix.
Related work that reduce the total training runtime, such as delayed gradient averaging~\cite{zhu2021delayed} and wait-free decentralized FL training~\cite{bornstein2022swift}, are designed for conventional FL and does not deal with the PFL+ scenario. 
Furthermore, we also provide a discussion on how our wait-free local training protocol could be adapted to the partial client participation scheme in FL in the Appendix.
 
\begin{algorithm}[t!]
  \caption{Spectral Co-Distillation with Wait-free Training for PFL+ }
  \label{alg: scd}
  \textbf{Inputs:} $N$, $T$, $\eta_\text{p}$, $\eta_\text{G}$, $w_\text{G}^{0}$, \{$w_{\text{p}, i}^{0}$\}$_{i=0}^N$, $E_\text{G}$, $E_\text{p}$ \\ 
  \textbf{Outputs:} Generic model $w_\text{G}^{T}$, personalized models \{$w_{\text{p}, i}^{T}$\}$_{i=1}^N$ \\[-0.9em]
  \begin{algorithmic}[1] 
    \FOR {$t=1$ \textbf{to} $T$}   
    \FOR{ each client $k=1$ \textbf{to} $N$ \textbf{in parallel}}
    \STATEx{\textit{\qquad // Generic model training and update} }
    \State{$w_{\text{G}, k}^{t} \gets$ \textsc{GmUpdate($w_{\text{G}}^{t-1},w_{\text{p},k}^{t-1}$)} }
     \State{Upload weights $w_{\text{G}, k}^{(t)}$ to server}
     \STATEx{\textit{\qquad // Personalized model training} \textcolor{blue}{(\texttt{Task 1}: Line 5)} }
    \State{$w_{\text{p}, k}^{t} \gets$ \textsc{PmUpdate}($w_{\text{p},k}^{t-1},w_{\text{G},k}^{t}$)            
    \State{\textbf{do sequentially} \textcolor{red}{(\texttt{Task 2}: Lines 7--8)}}
       \hspace{7em}\smash{$\left.\rule{0pt}{2.5\baselineskip}\right\}\ \mbox{Perform \texttt{Tasks \textcolor{blue}{1}} \& \texttt{\textcolor{red}{2}} in parallel}$} 
    \State{\quad Uplink communication of generic model to server }
    \State{\quad Generic model aggregation to obtain $w_{\text{G}}^{t}$} at server}
    \ENDFOR
\ENDFOR
\RETURN { $w_\text{G}^{T}$, \{$w_{\text{p},i}^{T}$\}$_{i=1}^N$}
\end{algorithmic}

\begin{algorithmic}[1]
\REQUIRE{\textsc{GmUpdate}}($w_{\text{G}}^{t-1},w_{\text{p},k}^{t-1}$)
\REQ{$w_{\text{G}}^{t-1},w_{\text{p},k}^{t-1}$ are the latest generic model and personalized model.}
\STATE{$w_0 \gets w_{\text{G}}^{t-1}$}
\FOR{$j=1$ \textbf{to} $E_\text{G}$}
\STATE{$w_j \gets w_{j-1} -\eta_\text{G} \nabla \mathcal{L}^{\text{G}} (w_{j-1}|w_{\text{p},k}^{t-1})$ \textit{\quad // Using truncated low frequency spectrum information} }
\ENDFOR
\RETURN {  $w_j$}
\end{algorithmic}

\begin{algorithmic}[1]
\REQUIRE{\textsc{PmUpdate}}($w_{\text{p},k}^{t-1},w_{\text{G},k}^{t}$)
\REQ{$w_{\text{p},k}^{t-1},w_{\text{G},k}^{t}$ are the latest personalized model and updated generic model of client $k$.}
\STATE{$w_0 \gets w_{\text{p},k}^{t-1}$}
\FOR{$j=1$ \textbf{to} $E_\text{p}$}
 \STATE{$w_j \gets w_{j-1} -\eta_\text{p} \nabla \mathcal{L}^{\text{p}} (w_{j-1}|w_{\text{G},k}^{t})$ \textit{\quad // Using full model spectrum information}}
\ENDFOR
\RETURN {  $w_j$}
\end{algorithmic}

\end{algorithm}

\subsection{Algorithm Summary}\label{subsec:algo}
Our proposed spectral co-distillation framework combined with our wait-free local training protocol, is given in \cref{alg: scd}.
As an overview, we begin every communication round $t$ with the server broadcasting the global generic model $w_{\text{G}}^{t-1}$ to each client for local computation. Each client $i$ would send back the updated generic model $w_{\text{G}, i}^{t}$ after $E_\text{G}$ local computation steps for global model aggregation, then immediately start the personalized model training and continue until the global generic model $w_{\text{G}}^{t}$ is received, which marks the start of the next communication round $t+1$.

\textbf{Remark on convergence analysis.} 
Note that the global loss function includes a weighted sum of the local loss functions and a regularizer. The regularizer is given in the form of the divergence function $\mathfrak{D}$, which is equivalent to KL divergence; cf. Sec. \ref{subsec:pm_train}. As demonstrated in \cite{akbari2021does}, the KL divergence usually exhibits convexity in terms of the model parameters. Consequently, since the model training undergoes (stochastic) gradient descent, it is possible to establish a convergence rate for the training of the global model (under the commonly employed assumption of smoothness of the local loss functions).

\renewcommand\arraystretch{1.2}
\begin{table}[t!]
\centering
\begin{adjustbox}{width=1\textwidth,center}
\begin{tabular}{l|cc|cc|cc}
\hline
\multirow{2}{*}{Methods} & \multicolumn{2}{c|}{$\alpha=1$}       & \multicolumn{2}{c|}{$\alpha=0.5$}     & \multicolumn{2}{c}{$\alpha=0.1$}      \\ \cline{2-7} 
                         & \multicolumn{1}{c|}{GM} & PM & \multicolumn{1}{c|}{GM} & PM & \multicolumn{1}{c|}{GM} & PM \\ \hline
FedAvg            & 85.35 $\pm$ 0.11    & (80.33 $\pm$ 0.38)  & 80.76 $\pm$ 0.13   & (74.51 $\pm$ 0.48) & 73.51 $\pm$ 0.17 & (72.68 $\pm$ 0.39) \\
FedProx           & 85.61 $\pm$ 0.08    & (86.28 $\pm$ 0.21)  & 80.54 $\pm$ 0.14   & (76.88 $\pm$ 0.30) & 71.96 $\pm$ 0.12    & (73.77 $\pm$ 0.30) \\
FedDyn            & 86.03 $\pm$ 0.13    & (85.33 $\pm$ 0.19)  & 80.88 $\pm$ 0.18   & (78.93 $\pm$ 0.25) & 73.62 $\pm$ 0.14    & (74.25 $\pm$ 0.58) \\
FedGen   & 86.17 $\pm$ 0.32   & ( 85.24 $\pm$ 0.47)  & 79.86 $\pm$ 0.34   & (77.52 $\pm$ 0.43) & 71.36 $\pm$ 0.28    & (71.42 $\pm$ 0.63) \\
FedAvgM           & 85.44 $\pm$ 0.05    & (82.85 $\pm$ 0.28) &  81.04 $\pm$ 0.09   & (75.71 $\pm$ 0.33) & 72.87 $\pm$ 0.06   & (72.96 $\pm$ 0.14) \\ \hline
pFedMe            & 85.58 $\pm$ 0.23    & 88.17 $\pm$ 0.17  & 79.33 $\pm$ 0.14  & 84.66 $\pm$ 0.17 & 72.11  $\pm$ 0.23    &81.18 $\pm$ 0.15\\
Ditto             & 85.34 $\pm$ 0.10    & 87.55 $\pm$ 0.09  &   80.70 $\pm$ 0.13  &83.39 $\pm$ 0.12 &  73.45 $\pm$ 0.18    & 80.08 $\pm$ 0.20 \\
FedRep            & (85.61 $\pm$ 0.19)        & 87.32 $\pm$ 0.11  & (80.33 $\pm$0.23)  & 84.10 $\pm$ 0.10 & (73.50  $\pm$  0.24)   & 79.74 $\pm$ 0.31 \\
FedRoD            & 86.02 $\pm$ 0.12   & 91.67 $\pm$ 0.16  & \textbf{81.31  $\pm$ 0.15}  & 85.91 $\pm$ 0.15 &  74.64 $\pm$ 0.07    & 81.37 $\pm$ 0.17 \\
FedBABU           &  (85.67 $\pm$ 0.24)   & 91.34 $\pm$ 0.19  &  (79.57 $\pm$ 0.23)  & 83.22 $\pm$ 0.33 &  (73.88 $\pm$0.19)     &80.58 $\pm$ 0.22\\\hline
Ours              & \textbf{86.37 $\pm$ 0.15}   &\textbf{92.25 $\pm$ 0.21}  & 81.27 $\pm$ 0.18  &\textbf{86.59 $\pm$ 0.17} &  \textbf{75.52 $\pm$ 0.11}    & \textbf{82.69 $\pm$ 0.16} \\ \hline
\end{tabular}
\end{adjustbox}
\caption{Average (3 trials) and standard deviation of the best test accuracies for generic/personalized models of various methods on CIFAR-10 with different non-IID settings. See also Remark \ref{remark}.}
\label{tab:acc_cifar10}
\end{table}

\begin{table}[t!]
\centering

\begin{adjustbox}{width=0.83\textwidth,center}
\begin{tabular}{l|cc|cc}
\hline
\multirow{2}{*}{Methods} & \multicolumn{2}{c|}{$\alpha=1$}           & \multicolumn{2}{c}{$\alpha=0.1$}      \\ \cline{2-5} 
                         & \multicolumn{1}{c|}{GM} & PM & \multicolumn{1}{c|}{GM} & PM \\ \hline
FedAvg                   &  48.37 $\pm$ 0.22   &   (52.64 $\pm$ 0.48)    &   38.61 $\pm$ 0.27   & (39.27 $\pm$ 0.42) \\
FedProx                  &  47.33 $\pm$ 0.15   &   (53.85 $\pm$ 0.33)    &   39.55 $\pm$ 0.18   & (41.33 $\pm$ 0.38) \\
FedDyn                   &  49.24  $\pm$ 0.27  &   (57.20 $\pm$ 0.35)    &   40.43 $\pm$ 0.14   & (40.92 $\pm$ 0.26) \\
FedAvgM                  &  48.55  $\pm$ 0.19  &   (55.60 $\pm$ 0.26)    &   39.03 $\pm$ 0.08   & (40.85 $\pm$ 0.19) \\ \hline
pFedMe                   &  47.29 $\pm$ 0.27   & 61.52 $\pm$ 0.25     & 38.22  $\pm$ 0.23       &  45.88 $\pm$ 0.32\\
Ditto                    &  48.37 $\pm$ 0.25   & 60.47 $\pm$ 0.27      & 39.61 $\pm$ 0.19         &  43.12 $\pm$ 0.28 \\
FedRep                   &  (46.32  $\pm$ 0.23) & 58.76 $\pm$ 0.36   &  (40.11$\pm$ 0.35)        &  45.22 $\pm$ 0.19\\
FedRoD                   & 50.07 $\pm$ 0.16    &  62.51 $\pm$ 0.15     &    40.58 $\pm$ 0.22      & 45.99 $\pm$ 0.14   \\
FedBABU                  &  (48.52  $\pm$ 0.30)  &  60.33 $\pm$ 0.28    &  (37.35  $\pm$ 0.29)       &  44.72 $\pm$ 0.28 \\ \hline
Ours                     & \textbf{51.39 $\pm$ 0.22}    &  \textbf{63.15 $\pm$ 0.16}     &   \textbf{40.67 $\pm$ 0.14}       &\textbf{46.82 $\pm$ 0.23}   \\ \hline
\end{tabular}
\end{adjustbox}
\caption{Average (3 trials) and standard deviation of the best test accuracies for generic/personalized models of various methods on CIFAR-100 with different non-IID settings. See also Remark \ref{remark}.}
\label{tab:acc_cifar100}
\end{table}

\section{Experiments}\label{sec:exp}
\subsection{Experiment setup}
\textbf{Datasets, DNN models, federated settings, and evaluation metrics.} We evaluated our proposed PFL+ framework with $N$ clients on CIFAR-10/100~\cite{krizhevsky2009learning}, and iNaturalist-2017, using model architectures ResNet-18/34~\cite{he2016deep} and ResNet-50, respectively. 
For the experiments on CIFAR-10 (resp. CIFAR-100), we used $N=100$ (resp. $N=50$).
For experiments on iNaturalist-2017~\cite{van2018inaturalist}, we used $N=20$. 
For dataset partition, we used the symmetric Dirichlet distribution to emulate real-world heterogeneous data distributions~\cite{hsu2019measuring,acar2020federated}, where the heterogeneity is controlled by the concentration parameter $\alpha$. (A smaller $\alpha$ indicates a higher degree of data heterogeneity.)
For evaluation, we used two performance metrics:  
\begin{itemize}
    \item Generic model evaluation: global test accuracy (same metric in conventional FL).
    \item Personalized model evaluation: weighted average of local test accuracies.
\end{itemize}
For every client, the PM is evaluated on a local test set, whose underlying distribution is the same as that for the local training set.
All the experiments are implemented with a full client participation scheme. Further experiment details, results on partial client participation, and the computation overhead discussion are provided in the Appendix. 

\begin{remark}\label{remark}
For generic FL methods, personalized model (PM) accuracies are obtained by evaluating the generic model (GM) on local test sets. For PFL methods without GM training, GM accuracies are obtained by evaluating the averaged PM on the global test set.
\end{remark}

\textbf{Baselines.} We compared our proposed method with the following SOTA PFL methods: pFedMe~\cite{t2020moreau}, Ditto~\cite{li2021ditto}, FedRoD~\cite{achituve2021gpkernel}, FedRep~\cite{collins2021exploiting}, and FedBABU~\cite{oh2022fedbabu}.
Moreover, to have a fair performance evaluation of the generic models, we also include methods designed for conventional FL as baselines: FedAvg~\cite{mcmahan2017communication}, FedProx~\cite{li2020prox}, FedDyn~\cite{acar2020federated}, 
FedGen\cite{zhu2021data}, and FedAvgM~\cite{hsu2019measuring}.

\begin{table}[t!]
\begin{adjustbox}{width=0.98\textwidth,center}
\begin{tabular}{c|cccccccc}
\hline
Methods   & FedProx & FedDyn & Ditto    & FedRep & FedRoD & FedBABU     & Ours \\ \hline
GM        & 39.46$\pm$0.39 & 39.35$\pm$0.27   & 39.33$\pm$0.33   &  39.81$\pm$0.41  & 40.16$\pm$0.35   & 39.23$\pm$0.53 & \textbf{41.75$\pm$0.37}   \\ 
PM        & 41.58$\pm$0.27 & 40.99$\pm$0.35  & 41.88$\pm$0.41    &  42.07$\pm$0.24  & 44.54$\pm$0.29   &  42.36$\pm$0.44  & \textbf{45.87$\pm$0.21}     \\ \hline
\end{tabular}
\end{adjustbox}
\caption{Average (3 trials) and standard deviation of the best test accuracies for generic/personalized models of various methods on iNaturalist-2017 with non-IID setting $\alpha=0.1$. See also Remark \ref{remark}.}
\label{tab:inat}
\end{table}

\subsection{Performance comparison with state-of-the-art methods}
We evaluated the generalizability of our proposed spectral co-distillation framework, as well as the communication cost performance of our wait-free training protocol for PFL+. 

\textbf{Generalizability over heterogeneous settings.}
We compared the best test accuracies with multiple baselines over the different levels of data heterogeneity, using the same system configuration. \cref{tab:acc_cifar10} and \cref{tab:acc_cifar100} give the main results on CIFAR-10 and CIFAR-100, respectively. In summary, our proposed framework achieves the best test accuracies across diverse heterogeneous data settings, outperforming all PFL and conventional FL baselines on both PM and GM test accuracies concurrently. 
We also investigated the performance on the real-world dataset iNaturalist2017 in \cref{tab:inat}, where our proposed method also achieves the best GM/PM test accuracies.
We attribute such consistent outperformance to the bi-directional co-distillation design. 
\begin{wraptable}{r}{0.62\textwidth}
\resizebox{\linewidth}{!}{
\begin{tabular}{l|cccc}
\hline
\multirow{3}{*}{Methods} & \multicolumn{2}{c|}{3 epochs}    & \multicolumn{2}{c}{5 epochs} \\ \cline{2-5} 
                         & 40\% & \multicolumn{1}{c|}{80\%} & 40\%          & 80\%         \\ \cline{2-5} 
                         & \multicolumn{4}{c}{Speedup}                                     \\ \hline
  
Ours (w/ WF)  & 1.82 $\times$ & 1.56 $\times$   &  2.21$\times$       & 1.85$\times$\\ \hline
  
Ditto w/ WF &1.97$\times$ & 1.38 $\times$    & 2.87$\times$       & 1.93$\times$ \\ \hline
 
FedRoD w/ WF & 1.75$\times$& 1.54 $\times$   & 2.42$\times$       & 2.19$\times$ \\ \hline

\end{tabular}}
\caption{Communication cost comparison of various methods for personalized model accuracies on CIFAR-10 to reach target accuracy (40\%/80\%) with non-IID setting $\alpha=0.1$. The speedup factors are with respect to the performance of the corresponding methods without WF.}
\label{tab:comm_cost}
\end{wraptable}
This demonstrates that: a) the spectral information of the generic model is useful for knowledge distillation during personalized model training; and b) using truncated spectral information of the personalized models could boost the performance of the generic model via careful spectrum truncation. (See Appendix for a sensitivity analysis of the truncation ratio $\tau$ and other hyper-parameters.)

\textbf{Communication cost comparison.}  
To demonstrate the superiority of the wait-free training protocol (\textbf{WF}), we evaluated the communication cost performance of SOTA methods with/without the protocol on non-IID CIFAR-10 ($\alpha=0.1$), in terms of the total runtime $\zeta_\text{total}$ for PM to reach the target test accuracy (40\%/80\%). 
A smaller $\zeta_\text{total}$ indicates higher communication efficiency.  
For PFL methods that train generic and personalized models using the compute-and-wait local training protocol, we evaluated Ditto and FedRoD.  We conduct experiments with different numbers of epochs for local PM training (3 or 5 epochs). 
As shown in \cref{tab:comm_cost}, our proposed wait-free training protocol could significantly improve the efficiency of convergence time and has the potential to boost the time efficiency of PFL+ methods.

\begin{table}[h!]
\begin{adjustbox}{width=0.86\textwidth,center}
\begin{tabular}{l|cc|cc}
\hline
\multirow{2}{*}{Method} & \multicolumn{2}{c|}{$\alpha=1$}       & \multicolumn{2}{c}{$\alpha=0.1$}      \\ \cline{2-5} 
                    & \multicolumn{1}{c|}{GM} & PM & \multicolumn{1}{c|}{GM} & PM \\ \hline
Ours                    &  \textbf{86.37 $\pm$ 0.15}    &\textbf{92.25 $\pm$ 0.21}   &   \textbf{75.52 $\pm$ 0.11}    & \textbf{82.69 $\pm$ 0.16}    \\ \hline
Ours w/o SCD-GM     &  85.35 $\pm$ 0.11   & 91.86 $\pm$ 0.17 &  73.51 $\pm$ 0.17  & 81.03 $\pm$ 0.20   \\
Ours w/o SCD-PM     &  82.74 $\pm$ 0.39   & 79.65 $\pm$ 0.83 &  68.96 $\pm$ 0.47  & 70.51 $\pm$ 1.21  \\ 
Ours w/o Both       &  85.35 $\pm$ 0.11   & 79.65 $\pm$ 0.83  & 73.51 $\pm$ 0.17  & 70.51 $\pm$ 1.21   \\\hline
\end{tabular}
\end{adjustbox}
\caption{Ablation study results on non-IID CIFAR-10 (average and standard deviation of 3 trials). \textbf{SCD-GM} (resp. \textbf{SCD-PM}) represents the spectral distillation approaches adopted during the training of generic (resp. personalized) model.}
\label{tab:ablation}
\end{table}

\subsection{Ablation results}
\textbf{Ablation study.} In our proposed spectral co-distillation framework, we introduce the bi-directional spectrum knowledge distillation to bridge the training of generic and personalized models with the target for training good generic and personalized models simultaneously. 
To achieve the target, truncated and full model spectrum information are adopted in different training stages. Here, we conduct an ablation study to evaluate the effectiveness of these two components (see \cref{tab:ablation} for the effects of each component), in which we apply the distillation approaches in the two training stages separately.
In the setup where both SCD-PM and SCD-GM are removed (Case I), the GM training is identical to FedAvg. In the case of removing only SCD-PM while keeping SCD-GM (Case II), each PM would be trained locally without any knowledge distilled from the GM. This is akin to the client training its model by itself, separately from the server. Naturally, the PM performance would be drastically lower.
As SCD-GM is kept in Case II, where the GM is the student and the PM is the teacher, since the PM's performance is drastically lower, we would expect a drop in the GM's performance. Informally, the model would be worse off with the distillation of bad knowledge, than without distillation.

As demonstrated in \cref{tab:ablation}, both the distillation methods can boost the accuracy performance of generic and personalized models, whereas the bi-directional distillation can bridge the training performance of the generic and personalized models.
Specifically, we can observe that, the SCD-PM module effectively transfers the knowledge from the generic model to the personalized model and avoids over-fitting during local training. 

 \begin{wrapfigure}{r}{0.65\textwidth}
 
\centering\includegraphics[width=0.52\textwidth]{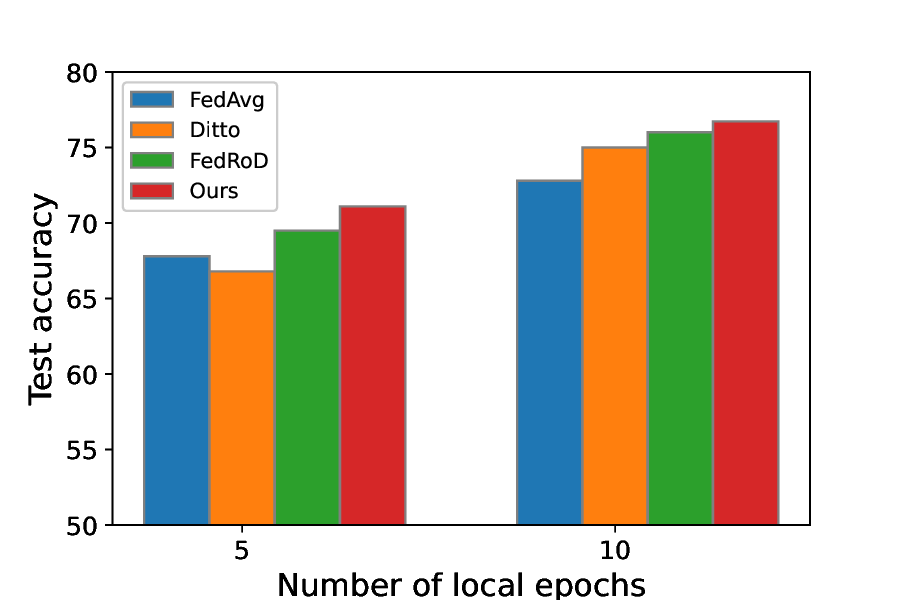} 
\caption{Performance comparison for generalizability on new clients of various methods.}
\label{fig:newclients}
\end{wrapfigure}

\textbf{Generalizability on new joining clients.}
In a real-world PFL system, dynamic client participation should be regarded as an important factor to consider during algorithm design, in which there would be continually new clients joining the system during training. 
The PFL system needs to rapidly train a good personalized model that could generalize well on the new client's local data.
To evaluate the generalizability of the system, we simulate a dynamic participation system with $80$ in-training clients and $20$ new clients on CIFAR-10 (partitioned by the Dirichlet distribution with $\alpha=0.1$), and deal with new clients with the global model-based fine-tuning approach. \cref{fig:newclients} gives the results of the average test accuracies of the new clients. Among all evaluated methods, our method has the best average test accuracies, illustrating the fast adaptive capability of our method.

\section{Conclusion}
In this work, we propose a spectral co-distillation framework for PFL to learn better generic and personalized models simultaneously. 
As far as we know, this is the first work in PFL that represents the (dis-)similarity of models via their Fourier spectra.
Even without co-distillation, there have been no other works that explore spectral distillation in PFL (or even in FL).
The advantage of this new approach is clear from our experiments: We achieved outperformance in both generic and personalized model training. 
Our framework also incorporates a simple yet effective wait-free local training protocol to reduce the overall local training time. 

\textbf{Limitations.} Our proposed spectral co-distillation framework, as currently formulated, does not deal with stragglers and adversarial attacks. Their influence on performance would require further investigation.
Also, our protocol assumes a synchronized network connection, which may not be practical for scenarios with large system/network heterogeneity.
Moreover, it would be good to consider a more
realistic local training protocol design that takes into account the issues of network/system heterogeneity; we leave the extension as future work.

\bibliographystyle{ieeetr}
\bibliography{scd.bib}

\end{document}